\documentclass{article} 
\usepackage{iclr2026_conference,times}

\usepackage{amsmath,amsfonts,bm}

\def\eqref#1{equation~\ref{#1}}

\def\1{\bm{1}}

\DeclareMathAlphabet{\mathsfit}{\encodingdefault}{\sfdefault}{m}{sl}
\SetMathAlphabet{\mathsfit}{bold}{\encodingdefault}{\sfdefault}{bx}{n}

\usepackage{graphicx}
\usepackage{amsmath }
\usepackage{enumitem}
\usepackage[utf8]{inputenc} 
\usepackage[T1]{fontenc}    
\usepackage{hyperref}       
\hypersetup{
    colorlinks=true,
    citecolor=BlueViolet,
    }
\usepackage{url}            
\usepackage{booktabs}       
\usepackage{amsfonts}       
\usepackage{nicefrac}       
\usepackage{microtype}      
\usepackage[dvipsnames]{xcolor}         
\usepackage{array}
\newcommand{\rotcol}[1]{\rotatebox{25}{\parbox{2.3cm}{ #1}}}

\usepackage{hyperref}

\title{RDAR: Reward-Driven Agent Relevance Estimation for Autonomous Driving}

\author{
  Carlo~Bosio\thanks{Work completed while the author was an intern at Zoox, Inc.}\\
  UC Berkeley\\
  \texttt{c.bosio@berkeley.edu} \\
  \And
  Greg~Woelki \\
  Zoox Inc. \\
  \texttt{gwoelki@zoox.com} \\
  \And
  Noureldin~Hendy \\
  Zoox Inc. \\
  \texttt{nhendy@zoox.com} \\
  \AND
  \hspace{3cm}Nicholas~Roy \\
  \hspace{3cm}Zoox Inc. \\
  \hspace{3cm}\texttt{nroy@zoox.com} \\
  \And
  Byungsoo~Kim \\
  Zoox Inc. \\
  \texttt{bykim@zoox.com} \\
}

\arxivcopy
\begin{document}

\maketitle

\begin{abstract}
Human drivers focus only on a handful of agents at any one time. On the other hand, autonomous driving systems process complex scenes with numerous agents, regardless of whether they are pedestrians on a crosswalk or vehicles parked on the side of the road.
While attention mechanisms offer an implicit way to reduce the input to the elements that affect decisions, existing attention mechanisms for capturing agent interactions are quadratic, and generally computationally expensive.
We propose RDAR, a strategy to learn per-agent relevance — how much each agent influences the behavior of the controlled vehicle — by identifying which agents can be excluded from the input to a pre-trained behavior model.
We formulate the masking procedure as a Markov Decision Process where the action consists of a binary mask indicating agent selection.
We evaluate RDAR on a large-scale driving dataset, and demonstrate its ability to learn an accurate numerical measure of relevance by achieving comparable driving performance, in terms of overall progress, safety and performance, while processing significantly fewer agents compared to a state of the art behavior model.
\end{abstract}

\section{Introduction}

Humans, when driving, do not pay equal attention to all agents around them (e.g., other vehicles, pedestrians). 
Transfomer-based attention models offer the promise of attending only to relevant components of the input, but existing attention models are typically quadratic in the size of the input space. Driving models encounter hundreds of input tokens, leading to substantial computational complexity and latency \cite{harmel2023scaling, huang2024drivegpt, baniodeh2025scaling}. 

In autonomous driving, there is a tension between the limited available compute resources and the desire to take advantage of scaling laws, large models, and test-time compute. Having access to numerical per-agent relevance scores would not only improve the interpretability of large driving models, but also allow compute resources to be prioritized for the features that are most important. 
In fact, when agents and other scene elements are represented explicitly as tokens, reasoning about interactions between these tokens (typically through self-attention or graph neural network operations) is quadratic and difficult to reduce using low-rank or other approximations that work well for long-sequence data. Reducing the number of tokens under consideration provides quadratic improvements in FLOPs used.
\begin{figure}[t]
    \centering
    \includegraphics[width=0.95\linewidth]{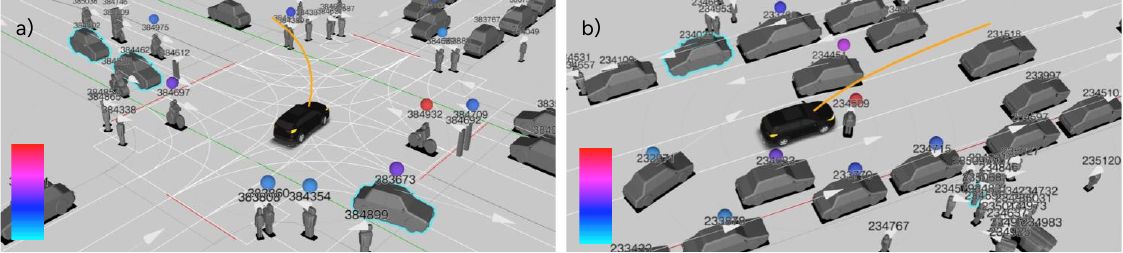}
    \caption{Example visualizations showing agent relevance assigned by our method. The top-$k$ relevant agents are labeled with a colored dot hovering over them. As shown by the scale in the bottom left of each image, red corresponds to higher relevance and light blue corresponds to lower relevance. a) attending to cyclist while turning left, b) attending to pedestrian during stop. The controlled vehicle is in black.}
    \label{fig:score-colorcode}
\end{figure}

In this work, we introduce RDAR (Reward-Driven Agent Relevance), through which we quantify agent relevance through a learned approach. The basic intuition is that if an agent is not relevant towards the driving decisions of the controlled vehicle, then its absence would not change the controlled vehicle's driving behavior significantly. Thus, we quantify per-agent relevance by learning which agents can be masked out from the controlled vehicle's planner input while maintaining a good driving behavior. 
We formulate agent selection as a reinforcement learning (RL) problem where an action is a binary mask indicating which agents to include in the driving policy input, and which not to. 
The RDAR scoring policy is trained in the loop with a frozen, pre-trained driving policy and a simulator. At each time step, based on the relevance scores, an agent mask is fed to the driving policy, making the controlled vehicle blind to the lower score agents. As it will be clear from the following sections, this is not a binary classification problem over agents due to the underlying system dynamics (e.g., not observing an agent now could lead to a collision later) and to the unavailability of ground truth labels.
Some examples of relevance (color-coded) computed by our method are shown in Fig. \ref{fig:score-colorcode}.
Our main contributions are:
%
\begin{itemize}[leftmargin=*]
\itemsep 0.0cm
    \item A novel reinforcement learning formulation for agent relevance estimation;
    \item A sampling-based mechanism for agent selection that enables efficient training and inference;
    \item A comprehensive evaluation showing  that we can maintain driving performance while processing only a handful of surrounding agents.
\end{itemize} 
\section{Related work}
Learning object ranking is a long standing problem in deep learning and typically requires human-labeled data \cite{cohen1997learning, burges2005learning,jamieson2011active}. In fields such autonomous driving, a manual ranking process can be not straightforward, and require large amounts of labeled data.
Input attribution \cite{sundararajan2017axiomatic} is a family of post-hoc analysis methods attempting to pinpoint which parts of the input are most responsible for a prediction.
Attribution methods mostly focus on leave-one-out (LOO) approaches \cite{liu2024attribot}, where chunks of the input are individually removed, or masked, and are correlated with changes in model outputs. 

Ranking agents in a driving scene based on their relevance is useful for both offline and online applications. 
Current autonomous driving systems quantify the relevance of surrounding agents either through fast, heuristic-based modules (e.g. based on euclidean distance), or learned models trained through supervised learning. Some approaches have been proposed for the supervision of these models, and they predominantly focus on LOO strategies coming from the attribution literature for agent prioritization \cite{refaat2019agent}, selective prediction \cite{ tolstaya2021identifying}, or offline introspection \cite{cusumano2025robust}.
While these LOO approaches provide insights into individual agent contributions, they have some limitations. First of all, a change in driving behavior, captured by a shift in predicted action, represents a \textit{different} driving behavior, but not necessarily a \textit{worse} one.  Second, these methods require multiple forward passes through a model (a planner in this case). Third, they do not capture the temporal dependencies caused by system dynamics. In this work, we propose a reward-driven method trained through reinforcement learning to estimate agent relevance.
%

RDAR computes per-agent relevance through just one forward pass, and is reward-driven instead of supervised through ground truth labels.
%
Reinforcement learning (RL) approaches have shown promising results in autonomous driving applications \cite{kiran2021deep}, and are being integrated in production and deployed in real-world systems. Several works have shown large scale urban driving through the use of both RL and human-collected real world driving data \cite{harmel2023scaling, gulino2023waymax, lu2023imitation}. In this work we build upon an existing learned behavior model, and train a scoring policy with closed-loop RL through a novel formulation for agent selection.
%
\begin{figure}[tb]
    \centering
    \includegraphics[width=0.6\linewidth]{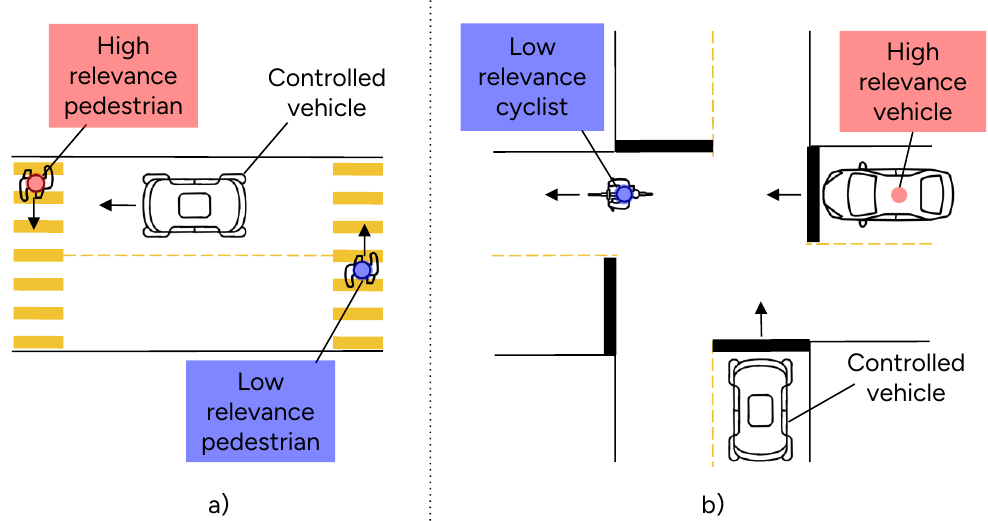}
    \caption{Example driving scenarios highlighting agent relevance. In a), the pedestrian crossing in front of the controlled vehicle is highly relevant, while the one behind is not. In b), the bike just went through the intersection and is not relevant anymore, while the car inching into the intersection is highly relevant.}
    \label{fig:examples-intro}
\end{figure}

\section{Problem Setup\label{sec:prob-setup}}
We wish to learn a policy $\pi^R_\theta$ assigning a \textit{relevance score} to each agent in the driving scene based on its influence on the driving behavior of the controlled vehicle ($\theta$ denotes learnable parameters).
We assume a pre-trained driving policy $\pi^D$ mapping scene information to driving actions is available. We also assume that, associated with the policy $\pi^D$, there is a reward function $r$ encoding some notion of good driving behavior. 
The RDAR scoring policy $\pi^R_\theta$ is trained in closed loop with the (frozen) driving policy $\pi^D$ and a driving simulator.

Formally defining a notion of agent relevance is not straightforward. 
However, human drivers have a good intuitive concept of such notion, which allows them to pay selective attention to surrounding agents. With reference to Fig. \ref{fig:examples-intro}a, a pedestrian crossing in front of the controlled vehicle is a highly relevant agent, because its presence its presence means that the controlled vehicle must come to a stop and yield instead of driving through a crosswalk. At the same time, the pedestrian crossing behind the driver has low relevance. Similar considerations are true for the intersection scenario of Fig. \ref{fig:examples-intro}b. The vehicle inching into the intersection has high relevance, because its presence means that the controlled vehicle must stop and yield. Instead, the cyclist who just passed through the intersection should not affect the controlled vehicle behavior. Therefore, we can say that an agent is relevant if hypothetically removing it from a driving scenario would cause the controlled vehicle to have a different behavior. 
\begin{figure*}[tb]
    \centering
    \includegraphics[width=0.9\linewidth]{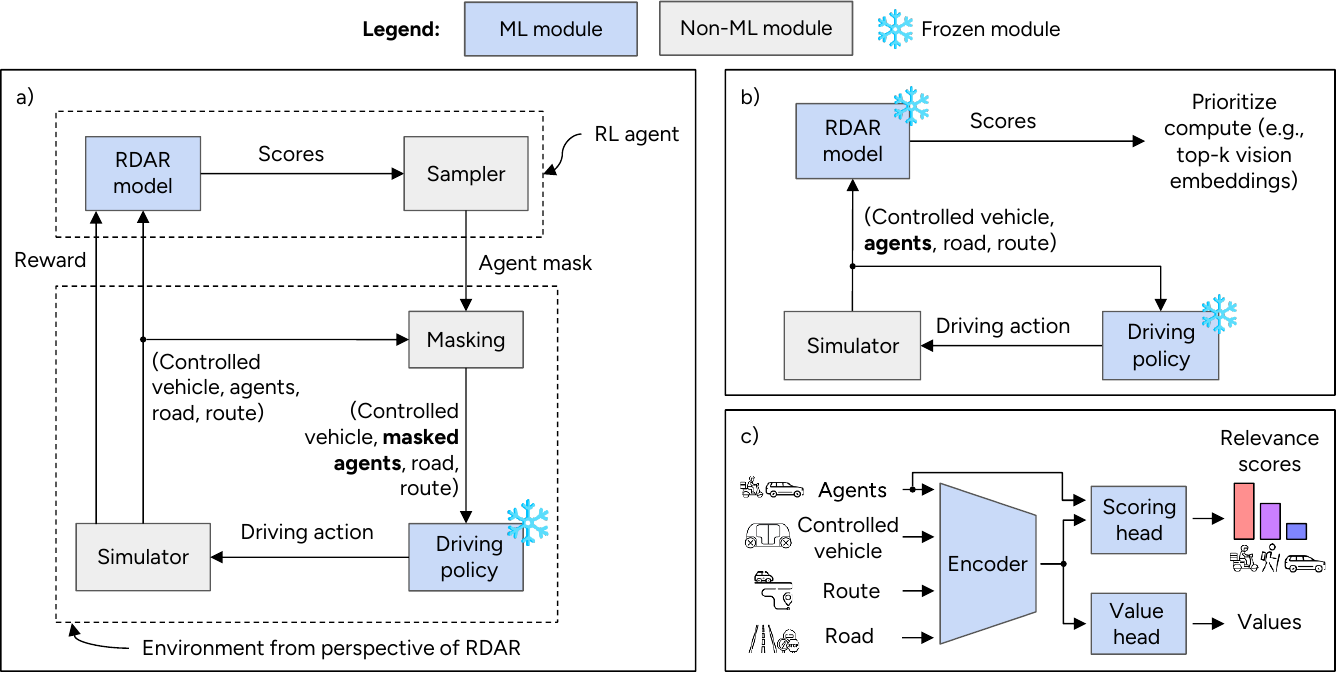}
    \caption{a) Block diagram of the system at training time. b) Block diagram of the system at deployment time. c) RDAR model structure.}
    \label{fig:block-diag}
\end{figure*}
%
\paragraph{Markov Decision Process formulation}
Following the above intuition,
we formulate the agent relevance estimation problem as a Markov Decision Process (MDP). The policy $\pi^R_{\theta}$ outputs per-agent relevance scores, which can be interpreted as logits of a categorical distribution. If agent $i$ is sampled from this \textit{relevance distribution}, it gets processed by the driving policy $\pi^D$, otherwise it is masked out and ignored by $\pi^D$. Given a hyperparameter $k\in\mathbb{N}$, an action is then a subset of $k$ surrounding agents, or a $k$-sample, to be processed by $\pi^D$. 
Our goal is thus to learn $\pi^R_\theta$ such that the return is maximized in expectation. Inaccurate relevance scores would make the driving policy blind to important agents in the scene, leading to low reward behaviors (e.g., collisions). The MDP setup for this process is defined by a standard tuple $(S, A, r, P, \mu_0)$, where:
\begin{itemize}[leftmargin=*]
\itemsep 0.0cm
    \item $S$ is the state space, including the controlled vehicle state, surrounding agent states (expressed in the controlled vehicle reference frame), road network and route information (see Fig.\ref{fig:block-diag}c);
    \item $A=\{0,1\}^N$ is the action space, consisting of binary masks of size $N$ (number of agents) indicating which agents to include in the planning input. The logits of the action distribution are the relevance scores (details in the following sections);
    \item $r$ is the reward function encoding good driving behavior. This is ideally the same reward or scoring function accompanying $\pi^D$;
    \item $P$ is the transition probability function associated to the environment. Note that the environment, from the perspective of the reinforcement learning agent $\pi^R_{\theta}$, consists of $\pi^D$ and the actual driving environment altogether (see Fig.\ref{fig:block-diag}a);
    \item $\mu_0$ is the initial state distribution.
\end{itemize}
The nature of the actions space makes this problem similar to a contextual multi-armed bandit (CMAB) \cite{lu2010contextual}, with the subtle difference that in this case the action changes the context, which in CMABs is assumed to be independent of the action.
\paragraph{Agent Selection Mechanism}
At each timestep $t$, the RDAR policy $\pi^R_\theta$ outputs a vector of per-agent relevance scores $\phi_t = [\phi_1, \phi_2, \ldots, \phi_N]$. At deployment time, the top-$k$ scoring agents are selected, while at training time, agents are randomly sampled to encourage exploration. For sampling, the scores are converted to a categorical distribution $p_t = [p_1, p_2, ..., p_N]$ over the binary agent selection action space through a softmax, where
\begin{equation}
    p_i = \frac{\exp(\phi_i)}{\sum_{j=1}^N \exp(\phi_j)}.
\end{equation}
Drawing one sample from this distribution corresponds to selecting one agent. If we draw exactly one sample, the probability of agent $i$ being selected is $p_i$. We can thus get a $k$-sample by drawing $k$ samples sequentially, without replacement, by renormalizing the probabilities at each step. 
%
We denote an agent $k$-sample as
\begin{equation}
    a=(a_1, a_2, \ldots, a_k), \,\,\,a_i\in\{1, ..., N\}, \label{eq:sample}
\end{equation}
where each component $a_i$ is the integer index corresponding to the selected agent. Note that this notation and an $N$-dimensional binary vector are equivalent. Then, the probability of selecting an ordered sample of agents without replacement is
\begin{equation}
    P(a_1, \ldots, a_k) = \prod_{i=1}^k \frac{p_{a_i}}{1 - \sum_{j=1}^{i-1} p_{a_j}},\label{eq:probs}
\end{equation}
where the denominators are the renormalization terms. Note that although we describe the sampling process as sequential, the Gumbel top-$k$ trick enables efficient, single-step sampling without replacement \cite{jang2016categorical, kool2019stochastic}. The trick consists of perturbing the distribution $p_t$ with a Gumbel distribution, and greadily selecting the top-$k$ elements: 
\begin{equation}
(a_1, ..., a_k)=\text{arg top-}k \,\,\left\{\,p_t - \log (-\log U)\,\right\},\,\,\,\, U\sim \text{Uniform}(0,1)
\end{equation}
By applying the logarithm to \eqref{eq:probs}, we can compute the log-likelihood of the $k$-sample:
\begin{equation}
    \log P(a_1, \ldots, a_k) = \sum_{i=1}^k \log p_{a_i} - \sum_{i=1}^k\log \Bigl(1 - \sum_{j=1}^{i-1} p_{a_j}\Bigr).
    \label{eq:log-likelihood}
\end{equation}
Since the scores are the output of our model $\pi^R_\theta$, \eqref{eq:log-likelihood} is exactly what we need for policy gradient updates in an RL framework. Once a $k$-sample is picked, only those $k$ agents are processed by the driving policy $\pi^D$. The action output by the driving policy is then applied to the simulator, and the overall state is updated. The simulator also produces the reward signal $r_t$ for RDAR. The process is then repeated.
\paragraph{Reinforcement learning framework}
We train the policy using an off-policy actor–critic framework with V-trace corrections \cite{espeholt2018impala}. The loss function combines four components:
\begin{equation}
    \mathcal{L} = \mathcal{L}\textsubscript{policy} + \lambda_c \mathcal{L}\textsubscript{critic} + \lambda_e \mathcal{L}\textsubscript{entropy} + \lambda_s \mathcal{L}\textsubscript{smoothing}. \label{eq:loss}
\end{equation}
The four components are respectively policy gradient loss, critic loss, entropy regularization loss, and action smoothing loss. Their exact expressions are:
\begin{align}
\mathcal{L}_{\text{policy}} &= - \mathbb{E}_t \left[ \rho_t \, \log \pi_\theta^R(a_t \mid s_t) \, \hat{A}^{\text{v-trace}}_t \right], \label{eq:pg-loss}\\
\mathcal{L}_{\text{critic}} &= \mathbb{E}_t \left[ \left( V_\theta(s_t) - V^{\text{target}}_t \right)^2 \right], \\
\mathcal{L}_{\text{entropy}} &= \mathbb{E}_t \left[ - \sum_{i=1}^N \pi_\theta^R(i \mid s_t) \log \pi_\theta^R(i \mid s_t) \right], \\
\mathcal{L}_{\text{smooth}} &= \mathbb{E}_t \left[\sum_{i=1}^N \| \pi_\theta^R(i \mid s_t) - \pi_\theta^R(i \mid s_{t-1}) \|^2 \right],
\end{align}
where $\rho_t = \frac{\pi_\theta^R(a_t \mid s_t)}{\mu(a_t \mid s_t)}$ are clipped importance weights, $a_t$ is the agent $k$-sample at time step~$t$ as in \eqref{eq:sample}. The $\hat{A}^{\text{v-trace}}_t$ and $V^{\text{target}}_t$ terms are computed following~\cite{espeholt2018impala}. The log-likelihood term in \eqref{eq:pg-loss} is computed as in \eqref{eq:log-likelihood}. The entropy and smoothing loss components, are calculated on the logits directly and do not depend on the $k$-sample $a_t$. The entropy component favors uniformity in the relevance scores, and therefore encourages exploration. The action smoothing component encourages the scores corresponding to the same agent to be consistent across time. It is possible that fewer than $N$ agents are physically present in a scene at a given time, in which case the loss terms corresponding to non-existing agents are masked. Finally, $\lambda_c$, $\lambda_e$, $\lambda_s$ are hyperparameters weighing the loss terms, selected empirically.

\section{Implementation}
\begin{figure*}[tb]
    \centering
    \includegraphics[width=0.99\linewidth]{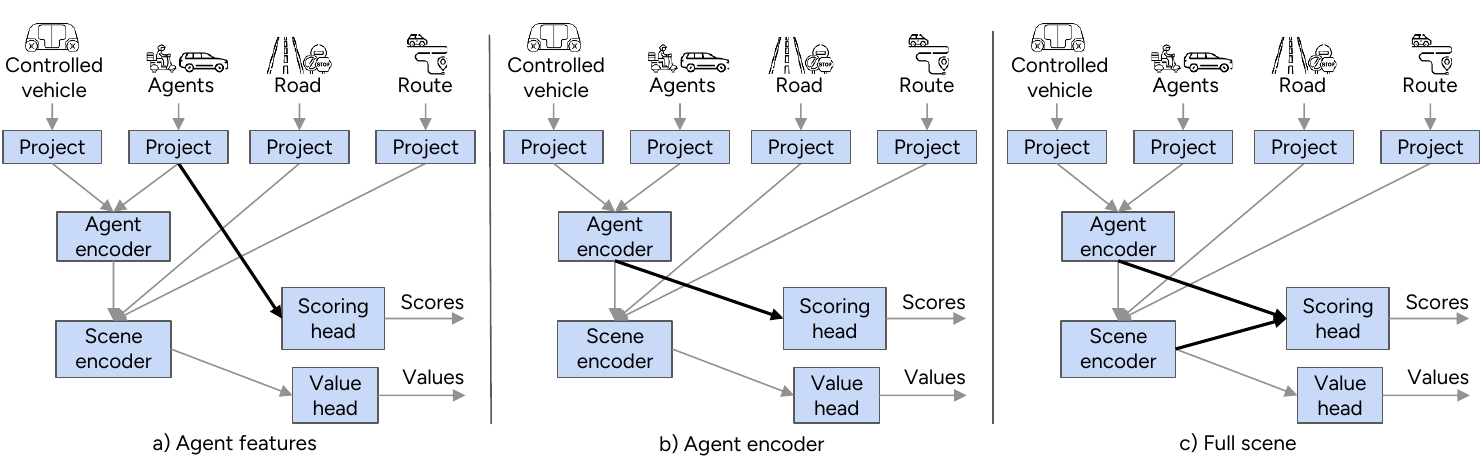}
    \caption{Different model architectures tested. Left: the scoring head consumes directly agent features projected through a linear layer. Middle: the scoring head consumes the output of the agent encoder, a transformer block cross attending among agents. Right: the scoring head consumes full scene information encoded into a latent embedding from the scene encoder, and attending back to agent tokens.}
    \label{fig:architectures}
\end{figure*}
\paragraph{Architecture}
The RDAR model architecture consists of three main components. An encoder module processes the full scene context (controlled vehicle state, surrounding agent states, road and route information) and computes embeddings. There are two heads: a scoring head, mapping embeddings to agent relevance scores, and a value head, approximating the value function. Note that the value function for the relevance scoring policy has a different meaning than the value function for the driving policy, since the expectation is over all possible $k$-samples rather than all possible driving actions.
The three options for the scoring head make use of embedding of varying depth from the encoder. The first option (Fig. \ref{fig:architectures}a) just uses the features from the agent projection layer. In this case, only agent state information is fed to the scoring head. The second option (Fig. \ref{fig:architectures}b) uses the embeddings output by the agent encoder module. In this case, the embeddings also encode agent interactions. The third option (Fig. \ref{fig:architectures}c) uses the output from the scene encoder, and reprojects it back to the agent level through a transformer block. In this case, all information from the driving scene is used to compute agent relevance.
The value head is kept the same for the three architectures, and uses all available information.
\paragraph{Training details}
The training-time agent sampling procedure described in Section \ref{sec:prob-setup} is done through the Gumbel top-$k$ trick \cite{kool2019stochastic}, natively used, for instance, in the JAX's implementation of the categorical distribution \cite{jax2018github}.
%
Also at training time, the number $k$ of agents sampled is randomized to make sure the model learns actual relevance scores and not only to differentiate between top-$k$ and non top-$k$ agents. We uniformly sample a different value $k$ per driving scenario.
%
To achieve scale, we use a distributed, asynchronous reinforcement learning infrastructure similar to IMPALA \cite{espeholt2018impala}. We found these hyperparameter values to give the best performances: learning rate $2\cdot 10^{-5}$, $\lambda_c=0.1$, $\lambda_e=0.2$, $\lambda_s=0.05$.
No sampling happens at deployment time, and and the top-$k$ agents are selected greadily (analogous to selecting the argmax action in standard RL). 

\section{Experimental Setup\label{sec:exp-setup}}
\begin{figure}[tb]
    \centering
    \includegraphics[width=0.9\linewidth]{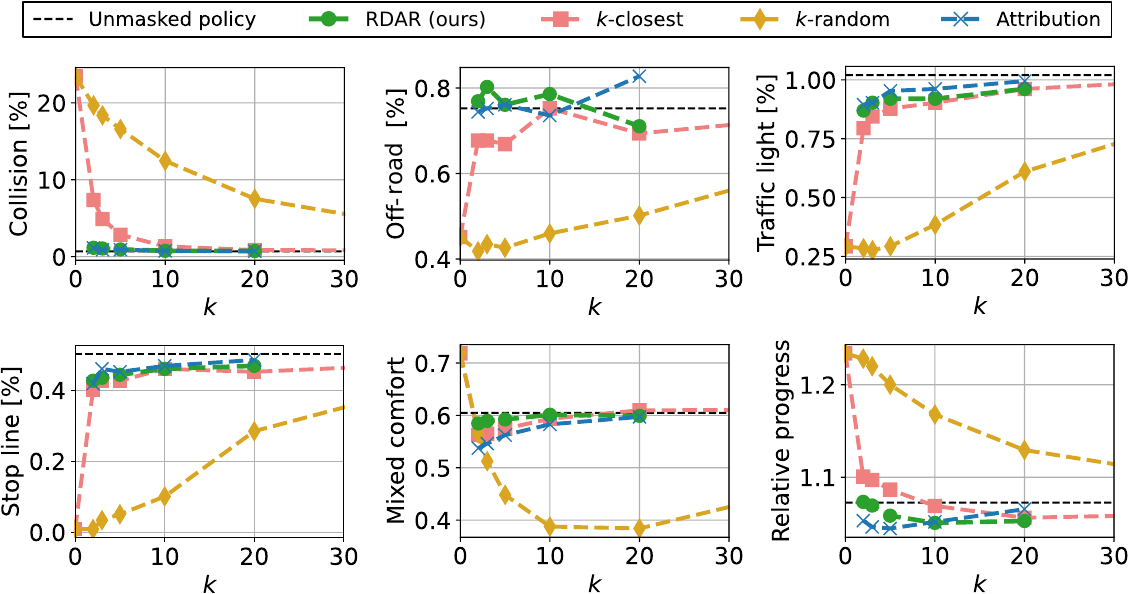}
    \caption{Metrics of best RDAR model and baselines against number $k$ of selected agents in closed-loop. All the closed-loop metrics for RDAR are close to the full policy baseline, supporting the claim that we are able to learn true agent relevance. In this case, the RDAR model is the version processing full scene information (see Fig. \ref{fig:architectures}c).}
    \label{fig:baselines}
\end{figure}

\paragraph{Datasets}
We train and evaluate our approach on large-scale proprietary datasets  consisting of ten-second long scenarios of real world, diverse urban driving. The training dataset contains around two million scenarios, while the evaluation dataset contains twelve thousand. The data were collected across Las Vegas (LV), Seattle
(SEA), San Francisco (SF), and the campus of the Stanford Linear Accelerator Center (SLAC). Note
that Las Vegas, Seattle, and San Francisco represent dense urban settings, while SLAC is a suburban
academic campus. Training and validation data are not separated by location, so a given validation
example may have training examples from the same location at other times. Scenarios are sampled
at $5\,\mathrm{Hz}$ from real-world driving logs.
%
\paragraph{Metrics}
To quantitatively evaluate our method, we use standard driving metrics which we compute on the 12k scenario evaluation set. In these evaluations, the candidate relevance scoring policies and baselines are used in closed-loop as filters, with the driving policy $\pi^D$ processing only a subset $k$ of all the agents present at any one time. At evaluation, we select the top-$k$ scoring agents greedily (analogously to selecting an action through argmax in standard RL). We use the following metrics:
%
\begin{enumerate}[leftmargin=*]
\itemsep 0.0cm
    \item \textit{Collisions [\%]}: percentage of scenarios in which a collision occurs (lower is better);
    \item \textit{Traffic light [\%]}: percentage of scenarios in which a traffic light is violated (lower is better);
    \item \textit{Stop line [\%]}: percentage of scenarios in which a stop line is skipped (lower is better);
    \item \textit{Off-road [\%]}: percentage of scenarios in which the vehicle drives off-road (lower is better);
    \item \textit{Comfort}: metric combining four motion aspects -- forward/backward acceleration, turning acceleration, and how suddenly or abruptly these accelerations change. These are weighted, averaged, then converted to a 0-1 score where 1 means smooth driving and 0 means jerky, uncomfortable motion (higher is better).
    \item \textit{Progress ratio}: relative progress along the route with respect to the ground truth human log;
    \item \textit{Complexity}: computation required by the scoring method as a function of the number $N$ of agents in a scene.
\end{enumerate}
%
%
\paragraph{Baselines}
We compare RDAR to other scoring strategies. The evaluation is done in closed-loop, using these strategies to pick the top-$k$ agents to be processed by $\pi^D$:
\begin{itemize}[leftmargin=*]
\itemsep 0.0cm
    \item Closest-$k$ selection: Select the $k$ closest agents to the controlled vehicle -- equivalent to agents' relevance scores being inversely proportional to their distance to the controlled vehicle;
    \item Random-$k$ selection: Randomly select $k$ agents from the scene -- equivalent to agents' relevance scores being drawn uniformly at random;
    \item Attribution-based scoring: scores obtained via input attribution \cite{cusumano2025robust}. At each timestep, the procedure is as follows: the pre-trained driving policy $\pi^D$ is evaluated  $N$ + 1 times (one with full scene information and one with each individual agent omitted in turn). For each agent, the Jensen–Shannon divergence between the action from its masked-out pass and the nominal full-scene pass is computed.
\end{itemize}
%
The overall performance of these baselines when varying $k$ is shown in Fig. \ref{fig:baselines}.
When $k=N$ no agents are masked out from the driving policy.

\section{Results}
\paragraph{Quantitative results}
We report plots showing the trends of the reference metrics when varying the value $k$ in Fig. \ref{fig:baselines}. As expected, all the metrics seem to converge to the baseline's when increasing the number of agents processed by the driving policy, with diminishing returns. We can see that RDAR has comparable performances to the attribution method, and requires only a fraction of the computation time ($O(1)$ instead of $O(N)$). RDAR is also able to drive with a fraction of percent performance regressions compared to the nominal, full policy, while processing significantly fewer agents. In fact, an RDAR model used in closed-loop with $k\ge10$ already enables comparable driving performances to the nominal unmasked policy while consuming an order of magnitude fewer agents.
With increasing $N$, as expected the collision rates decrease, the comfort scores increase (due to reduced flickering of agent selection). However, traffic infraction rates seem to have an upward trend. This makes sense, as for instance driving on empty roads would make it very easy to avoid infractions, and driving in heavy traffic makes it harder due to the higher risk of collisions.

\begin{figure}[tb]
    \centering
    \includegraphics[width=0.95\linewidth]{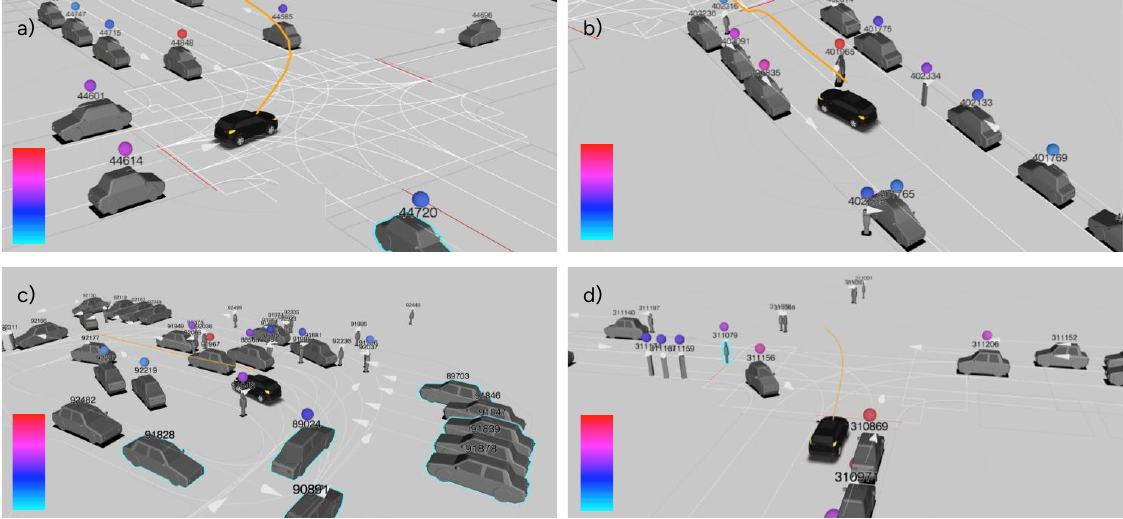}
    \caption{Example visualizations showing agent relevance assigned by our method. The top-$k$ relevant agents are labeled with a colored dot hovering over them. As shown by the scale in the bottom left of each image, red corresponds to higher relevance and light blue corresponds to lower relevance. The controlled vehicle is in black. Agent importance seems to be aligned with human intuition, highlighting high-risk pedestrians, other cars the controlled vehicle must yield to, agents in close proximity.}
    \label{fig:qualitative-res}
\end{figure}

Tab. \ref{tab:method_comparison} shows the actual numbers relative to the $k=10$ case. The three different architectures proposed in Fig. \ref{fig:architectures} show comparable performances when used in closed-loop with the driving policy. The model using full scene information (Fig. \ref{fig:architectures}c) leads to the least collisions and higher comfort score. There seems to be a trade-off between collision avoidance and the metrics related to traffic rules, which can be explained by the fact that avoiding collisions could require the controlled vehicle to do an infraction (e.g. driving off-road). In fact, while full scene information allows to minimize collisions, it does that at the cost of slightly higher infraction rates. Similarly, the baseline, unmasked driving policy has higher infraction rates than most masked alternatives. It is also interesting to see that the random scoring policy outperforms all the others when it comes to traffic rules (off-road, traffic lights, stop lines), which comes, as expected, at the cost of much higher collision rates. Scores computed using only agent features (Fig. \ref{fig:architectures}a) or attending to the controlled vehicle state (Fig. \ref{fig:architectures}b) achieves good closed-loop driving and requires fewer FLOPs compared to using full scene information (Fig. \ref{fig:architectures}c). 
On the other hand, full scene information enables enhanced situational awareness and lower collisions.
Therefore, the specific RDAR architecture can be chosen based on application needs in terms of memory, real-time compute requirements, or score accuracy.
\begin{table}[tb]
\setlength{\tabcolsep}{2pt}
\centering
\begin{tabular}{>{\arraybackslash}m{4.5cm}*{8}{>{\arraybackslash}m{1.1cm}}}
 & \rotcol{Collisions [\%]} & \rotcol{Off-road [\%]} & \rotcol{Traffic$\,\,$light [\%]} & \rotcol{Stop$\,\,$line [\%]} & \rotcol{Comfort} & \rotcol{Rel. progress} & \rotcol{Complexity} \\
\midrule
RDAR, Agent features (Fig. \ref{fig:architectures}a) & $0.94$ & $0.71$ & $0.97$ & $0.44$ & $0.57$ & $1.06$ & $O(1)$\\
RDAR, Agent encoder (Fig. \ref{fig:architectures}b) & $0.89$ & $0.70$ & $0.89$ & $0.43$ & $0.57$ & $1.06$ & $O(1)$\\
RDAR, Full scene (Fig. \ref{fig:architectures}c) & $0.77$ & $0.79$ & $0.92$ & $0.46$ & $0.60$ & $1.05$ & $O(1)$\\
\midrule
Closest-$k$ & $1.34$ & $0.75$ & $0.90$ & $0.46$ & $0.59$ & $1.07$ & $O(1)$\\
Random-$k$ & $12.5$ & $0.46$ & $0.38$ & $0.10$ & $0.39$ & $1.17$ & $O(1)$\\
Attribution & $0.75$ & $0.74$ & $0.96$ & $0.47$ & $0.58$ & $1.05$ & $O(N)$\\
\midrule
Baseline (no filter) & $0.68$ & $0.75$ & $1.02$ & $0.50$ & $0.61$ & $1.07$ & $\,\,\,-$\\
\end{tabular}
\caption{Closed-loop metrics of different methods with $k=10$ agents. Using only agent features or reasoning about agent interactions leads to good closed-loop driving performance, and requires fewer FLOPs compared to full scene, which on the other hand achieves lower collision rate.}
\label{tab:method_comparison}
\end{table}
%
\paragraph{Qualitative results}
We also report some visualizations from some closed-loop evaluation rollouts (the same used to compute the aggregate metrics) in Fig. 
\ref{fig:qualitative-res}. The scenes represented are challenging, cluttered driving scenes in which incorrect relevance quantification would lead to wrong masking and bad behaviors. In the figures, a colored dot hovers over the top-$10$ agents. These are the agents that are being processed by the driving policy, while the other are ignored. As shown in the color scale in the bottom left of each figure, the red color corresponds to high relevance score, the light blue color corresponds to lower relevance (still, within the top-$10$). These examples are from our best full-scene scoring policy (Fig. \ref{fig:architectures}c). We can see that the agent importance assigned by our model is aligned with human intuition, and highlights high-risk pedestrians, other cars the controlled vehicle must yield to, agents in close proximity.

\section{Discussion and Conclusions}
Our model provides insights into which agents influence driving decisions, enabling better understanding of planner behaviors. It can also inform how to allocate costly computation in a principled way--for example, running joint trajectory prediction or computing vision embeddings only for the most relevant agents.

This work opens several interesting directions. First, similar relevance-scoring methods could be applied to other components of the driving scene, such as road information. A challenge here is the potential for distribution shifts when masking inputs; in our case such effects are mild, since driving scenarios remain in-distribution regardless of the number or position of agents, but investigating mitigation strategies is important. Second, the scoring policy’s action space could be expanded beyond masking. Instead of excluding agents, the prioritization module could be trained to trigger targeted computation on selected agents, such as expensive vision embeddings, so that enhanced representations directly translate into downstream gains like improved driving rewards.

While our work is motivated by autonomous driving applications, the principle of learning per-agent relevance naturally extends to other multi-agent and multi-object settings, such as swarm robotics, multi-agent reinforcement learning, or video scene understanding, where only a subset of entities meaningfully influence a decision. By identifying and prioritizing relevant entities, similar approaches could improve scalability, reduce computational costs, and offer insights into the decision-making processes in these domains.
 
We introduced a reinforcement learning approach to estimate agent relevance in driving scenarios. By formulating relevance scoring as an agent-masking MDP, we enable end-to-end training of a scoring policy with a driving policy in the loop. Our method avoids costly post-hoc attribution and repeated forward passes, making it well suited for real-time autonomy stacks. In closed-loop evaluation, we show that comparable driving performance can be achieved while processing an order of magnitude fewer agents, highlighting the benefits of our approach in terms of behavior model introspection and dynamic compute allocation.



\bibliography{refs}
\bibliographystyle{iclr2026_conference}


\end{document}